\documentclass{llncs}
\usepackage{subfig}
\usepackage{url}
\usepackage{adjustbox}
\usepackage{float}
\usepackage{subfig}
\usepackage{cite}
\usepackage{multirow}
\usepackage{algorithm,algorithmic}
\usepackage{graphicx}
\usepackage{pgfplots}
\usepackage{tikz}
\usetikzlibrary{shapes,arrows}
\usetikzlibrary{positioning}
\usepackage{tkz-graph}
\usepackage{textcomp}
\usepackage{xcolor}
\usepackage{array}
\newcolumntype{C}[1]{>{\centering\let\newline\\\arraybackslash\hspace{0pt}}m{#1}}
\begin{document}
\title{\textit{Related Tasks can Share!} A Multi-task Framework for Affective language}
%
%
\author{Kumar Shikhar Deep \and Md Shad Akhtar \and Asif Ekbal \and Pushpak Bhattacharyya}
%
%
\institute{Department of Computer Science and Engineering, \\ Indian Institute of Technology Patna, India \\
\email{\{shikhar.mtcs17,shad.pcs15,asif,pb\}@iitp.ac.in}}

%
\maketitle              
\begin{abstract}
Expressing the polarity of sentiment as `\textit{positive}' and `\textit{negative}' usually have limited scope compared with the intensity/degree of polarity. These two tasks (i.e. sentiment classification and sentiment intensity prediction) are closely related and may offer assistance to each other during the learning process. In this paper, we propose to leverage the relatedness of multiple tasks in a multi-task learning framework. Our multi-task model is based on convolutional-Gated Recurrent Unit (GRU) framework, which is further assisted by a diverse hand-crafted feature set. Evaluation and analysis suggest that joint-learning of the related tasks in a multi-task framework can outperform each of the individual tasks in the single-task frameworks.    

\end{abstract}
\section{Introduction}
In general, people are always interested in what other people are thinking and what opinions they hold for a number of topics like product, politics, news, sports etc. The number of people expressing their opinions on various social media platforms such as Twitter, Facebook, LinkedIn etc. are being continuously growing. These social media platforms have made it possible for the researchers to gauge the public opinion on their topics of interest- and that too on demand. With the increase of contents on social media, the process of automation of Sentiment Analysis \cite{pang2002thumbs} is very much required and is in huge demand. User's opinions extracted from these social media platform are being used as inputs to assist in decision making for a number of applications such as businesses analysis, market research, stock market prediction etc. 

\begin{table}
\caption{Example sentences with their sentiment classes and intensity scores from SemEval-2018  dataset on Affect in Tweets \cite{mohammad2018semeval}.}
\label{fig:exm}
\centering
{
\begin{tabular}{|p{25em}|c|c|}
     \hline
     \bf Tweet & \bf Valence & \bf Intensity\\
     \hline \hline
     \textit{@LoveMyFFAJacket FaceTime - we can still annoy you} & \textit{Pos-S} & 0.677 \\ \hline
     \textit{and i shouldve cut them off the moment i started hurting myself over them} & \textit{Neg-M} & 0.283 \\ \hline 
     \textit{@VescioDiana You forgot \#laughter as well} & \textit{Pos-S} & 0.700 \\ \hline
\end{tabular}
}
\end{table}
Coarse-grained sentiment classification (i.e. classifying a text into either \textit{positive} or \textit{negative} sentiment) is a well-established and well-studied task \cite{kim2004determining}. However, such binary/ternary classification studies do not always reveal the exact state of human mind. We use language to communicate not only our sentiments but also the intensity of those sentiments, e.g. one could judge that we are very angry, slightly sad, very much elated, etc. through our utterances. Intensity refers to the degree of sentiment a person may express through his text. It also facilitates us to analyze the sentiment on much finer level rather than only expressing the polarity of the sentiments as positive or negative. In recent times, studies on the amount of positiveness and negativeness of a sentence (i.e. how positive/negative a sentence is or the degree of positiveness/negativeness) has gained attention due to its potential applications in various fields. Few example sentences are depicted in Table \ref{fig:exm}. 
In this work, we focus on the fine-grained sentiment analysis \cite{socher2013recursive}. Further, we aim to solve the fine-grained analysis with two different lenses i.e. fine-grained sentiment classification and sentiment intensity prediction.      
\begin{itemize}
    \item \textbf{Sentiment or Valence\footnote{Valence signifies the pleasant/unpleasant scenarios.} Classification:} In this task, we classify each tweet into one of the seven possible fine-grained classes -corresponding to various levels of positive and negative sentiment intensity- that best represents the mental state of the tweeter, i.e. very positive (\textbf{\em Pos-V}), moderately positive (\textbf{\em Pos-M}), slightly positive (\textbf{\em Pos-S}), neutral (\textbf{\em Neu}), slightly negative (\textbf{\em Neg-S}), moderately negative (\textbf{\em Neg-M}), and very negative (\textbf{\em Neg-V}).
    
    \item \textbf{Sentiment or Valence Intensity Prediction:} Unlike the discrete labels in the classification task, in intensity prediction, we determine the degree or arousal of sentiment that best represents the sentimental state of the user. The scores are a real-valued number in the range 0 \& 1, with 1 representing the highest intensity or arousal.
\end{itemize}

The two tasks i.e. sentiment classification and their intensity predictions are related and have inter-dependence on each other. Building separate system for each task is often less economical and more complex than a single multi-task system that handles both the tasks together. Further, joint-learning of two (or more) related tasks provides a great assistance to each other and also offers generalization of multiple tasks.

In this paper, we propose a hybrid neural network based multi-task learning framework for sentiment classification and intensity prediction for tweets. Our network utilizes bidirectional gated recurrent unit (Bi-GRU)\cite{Schuster:1997:BRN:2198065.2205129} network in cascade with convolutional neural network (CNN)\cite{lecun1995convolutional}. The max-pooled features and a diverse set of hand-crafted features are then concatenated, and subsequently fed to the task-specific softmax layer for the final prediction. We evaluate our approach on the benchmark dataset of SemEval-2018 shared task on Affect in Tweets \cite{mohammad2018semeval}. We observe that, our proposed multi-task framework attains better performance when both the tasks are learned jointly.

The rest of the paper are organized as follows. In Section \ref{sec:lit}, we furnish the related work. We present our proposed approach in Section \ref{sec:method}. In Section \ref{sec:exp}, we describe our experimental results and analysis. Finally, we conclude in Section \ref{sec:con}.

\section{Related Work} 
\label{sec:lit}
The motivation behind applying multi-task model for sentiment analysis comes from \cite{ruder2017overview} which gives a general overview of multi-task learning using deep learning techniques. Multitask learning (MTL) is not only applied to Natural Language Processing \cite{Collobert:2008:UAN:1390156.1390177} tasks, but it has also shown success in the areas of computer vision \cite{girshick2015fast},
drug discovery \cite{ramsundar2015massively} and many other.
The authors in \cite{duppada2018seernet} used stacking ensemble technique to merge the results of classifiers/regressors through which the handcrafted features were passed individually and finally fed those results to a meta classifier/regressor to produce the final prediction. This has reported to have achieved the state-of-the-art performance.

The authors in \cite{gee2018psyml} used bidirectional Long Short Term Memory (biLSTM) and LSTM with attention mechanism and performed transfer learning by first pre-training the LSTM networks on sentiment data. Later, the penultimate layers of these networks are concatenated to form a single vector which is fed as an input to the dense layers. There was a gated recurrent units (GRU) based model proposed by \cite{rozental2018amobee} with a convolution neural network (CNN) attention mechanism and training stacking-based ensembles. In \cite{meisheri2018tcs} they combined three different features generated using deep learning models and traditional methods in support vector machines (SVMs) to create an unified ensemble system. In \cite{park2018plusemo2vec} they used neural network model for extracting the features by transferring the emotional knowledge into it and passed these features through machine learning models like support vector regression (SVR) and logistic regression. In \cite{baziotis2018ntua} authors have used a Bi-LSTM in their architecture.  In order to improve the model performance they applied a multi-layer self attention mechanism in Bi-LSTM which is capable of  identifying salient words in tweets, as well as gain insight into the models making them more interpretable.

Our proposed model differs from previous models in the sense that we propose an end to end neural network based approach that performs both sentiment analysis and sentiment intensity prediction simultaneously. We use gated recurrent units (GRU) along with convolutional neural network (CNN) inspired by \cite{rozental2018amobee}. We fed the hidden states of GRU to CNN layer in order to get a fixed size vector representation of each sentence. We also use various features extracted from the pre-trained 
resources like DeepMoji\cite{felbo2017}, Skip-Thought Vectors\cite{kiros2015skip}, Unsupervised Sentiment Neuron\cite{radford2017learning} and EmoInt\cite{duppada2018seernet}.

 \begin{figure*}[!ht]
     \centering
     \includegraphics[width=\textwidth]{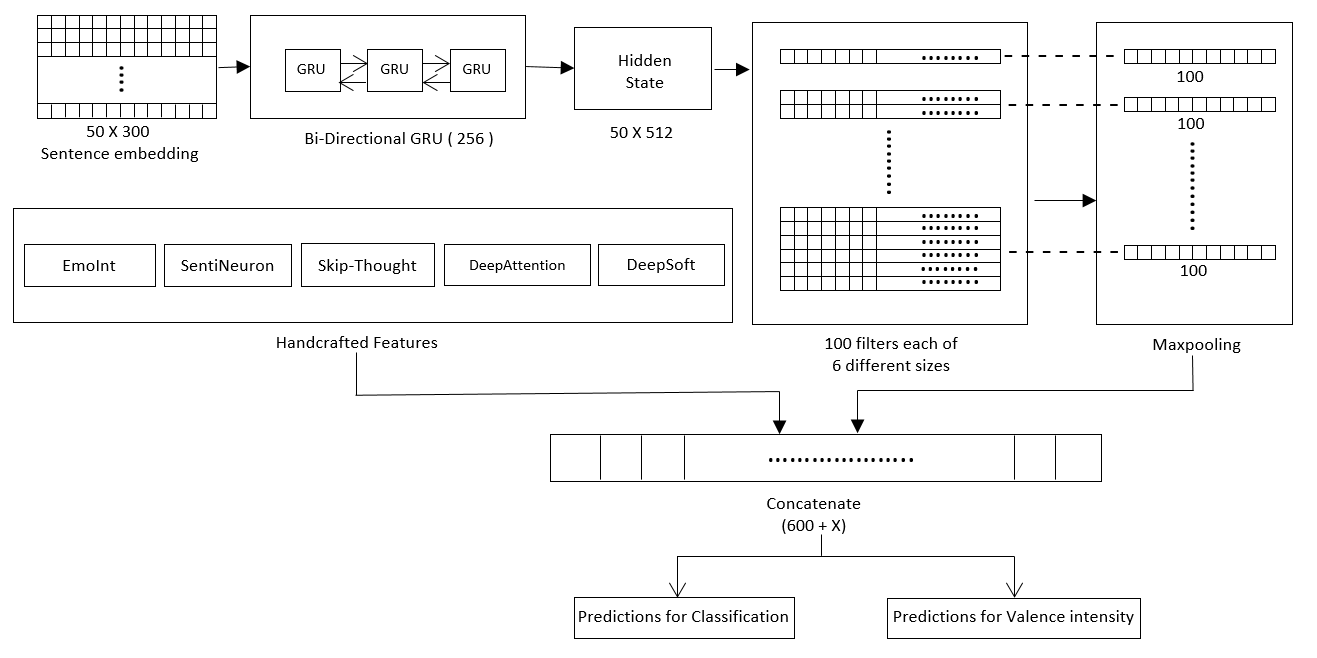}
     \caption{Proposed Architecture}
     \label{fig:my_label}
 \end{figure*}
 
\section{Proposed Methodology} \label{sec:method}
In this section, we describe our proposed multi-task framework in details. Our model consists of a recurrent layer (biGRU) followed by a CNN module. Given a tweet, the GRU learns contextual representation of each word in the sentence, i.e. the representation of each word is learnt based on the sequence of words in the sentence. This representation is then used as input to the CNN module for the sentence representation. Subsequently, we apply max-pooling over the the convoluted features of each filter and concatenated them. The hidden representation, as obtained from the CNN module, is shared across multiple tasks (here, two tasks i.e. sentiment classification and intensity prediction). Further, the hidden representation is assisted by a diverse set of hand-crafted features (c.f. section \ref{sec:feat}) for the final prediction. 

In our work, we experiment with two different paradigms of predictors i.e. a) the first model is the traditional deep learning framework that makes use of softmax (or sigmoid) function in the output layer, and b) the second model is developed by replacing softmax classifier using support vector machine (SVM) \cite{suykens1999least} (or support vector regressor (SVR)). 
 In the first model 
 we feed the concatenated representation to two separate fully-connected layers with softmax (classification) and sigmoid (intensity) functions for the two tasks. In the second model, we feed hidden representations as feature vectors to the SVM and SVR respectively, for the prediction. A high-level block diagram of the proposed methodology is depicted in Figure \ref{fig:exm}.    



\subsection{Hand-crafted Features}
\label{sec:feat}
We perform transfer learning from various state-of-the-art deep learning techniques. Following sub-sections explains these models in detail:
\begin{itemize}
    \item \textbf{DeepMoji} \cite{felbo2017}: DeepMoji performs distant supervision on a very large dataset \cite{thelwall2010sentiment} \cite{nakov2016semeval} (1.2 billion tweets) comprising of noisy labels (emojis). By incorporating transfer learning on various downstream tasks, they were able to outperform the state-of-the-art results of 8 benchmark datasets on 3 NLP tasks across 5 domains. Since our target task is closely related to this, we adapt this for our domain. We extract 2 different feature sets:
    \begin{itemize}
        \item the embeddings from the softmax layer which is of 64 dimensions.
        \item the embeddings from the attention layer which is of 2304 dimensions.
    \end{itemize}

    \item \textbf{Skip-Thought Vectors} \cite{kiros2015skip}: Skip-thought is a kind of model that is trained to reconstruct the surrounding sentences to map sentences that share semantic and syntactic properties into similar vectors. It has the capability to produce highly generic semantic representation of sentence. The skip-thought model has two parts:
    \begin{itemize}
        \item \textbf{Encoder : }It is generally a Gated Recurrent Unit (GRU) whose final hidden state is passed to the dense layers to get the fixed length vector representation of each sentence. 
        \item \textbf{Decoder : }It takes this vector representation as input and tries to generate the previous and next sentence. For this two different GRUs are needed.
    \end{itemize}
    Due to its fixed length vector representation, skip-thought could be helpful to us. The feature extracted from skip-thought model is of dimension 4800.

    \item \textbf{Unsupervised Sentiment Neuron}: \cite{radford2017learning} developed an unsupervised system which learned an excellent representation of sentiment. Actually the model was designed to produce Amazon product reviews, but the data scientists discovered that one single unit of network was able to give high predictions for sentiments of texts. It was able to classify the reviews as positive or negative, and its performance was found to be better than some popular models. They even got encouraging results on applying their model on the dataset of Yelp reviews and binary subset of the Stanford Sentiment Treebank. Thus the sentiment neuron model could be used to extract features by transfer learning. The features extracted from Sentiment Neuron model are of dimension 4096.
   
    \item \textbf{EmoInt} \cite{duppada2018seernet}: We intended to use various lexical features apart from using some pre-trained embeddings. EmoInt \cite{duppada2018seernet} is a package which provides a high level wrapper to combine various word embeddings. The lexical features includes the following:
    \begin{itemize}
        \item AFINN \cite{mohammad2017wassa} contains list of words which are manually rated for valence between -5 to +5 where -5 indicates very negative sentiment and +5 indicates very positive sentiment. 
        \item SentiWordNet \cite{baccianella2010sentiwordnet} is lexical resource for opinion mining. It assigns to each synset of WordNet three sentiment scores: positivity, negativity, objectivity.
        \item SentiStrength \cite{thelwall2010sentiment} gives estimation of strength of positivity and negativity of sentiment.
        \item NRC Hashtag Emotion Lexicon \cite{mohammad2017wassa} consists of emotion word associations computed via Hashtags on twitter texts labelled by emotions.
        \item NRC Word-Emotion Association Lexicon \cite{mohammad2017wassa} consists of 8 sense level associations (anger, fear, joy, sadness, anticipation, trust, disgust and surprise) and 2 sentiment level associations(positive and negative)
        \item The NRC Affect Intensity \cite{mohammad2017emotion} are the lexicons which provides real values of affect intensity. 
    \end{itemize}
    The final feature vector is the concatenation of all the individual features. This feature vector is of size (133, 1). 
 \end{itemize}

\subsection{Word Embeddings} \label{sec:we}
Embedding matrix is generated from the pre-processed text using a combination of three pre-trained embeddings:
\begin{enumerate}
    \item \textbf{Pre-trained GloVe embeddings for tweets} \cite{pennington2014glove}: We use 200-dimensional pre-trained GloVe word embeddings, trained on the Twitter corpus, for the experiments. To make it compatible with the other embeddings, we pad 100-dimensional zero vector to each embedding.
    \item \textbf{Emoji2Vec} \cite{eisner2016emoji2vec}: Emoji2Vec provides 300 dimension vectors for most commonly used emojis in twitter platform (in case any emoji is not replaced with its corresponding meaning).
    \item \textbf{Character-level embeddings}\footnote{\url{https://github.com/minimaxir/char-embeddings}}: Character-level embeddings are trained over common crawl glove corpus providing 300 dimensional vectors for each character (used in case if word is not present in other two embeddings).
    
\end{enumerate}
 Procedure to generate representations for a tweet using all these embeddings is described in Algorithm \ref{algo}. 
\begin{algorithm}
\begin{algorithmic}
\FOR{\textit{word} in \textit{tweet}}
  \IF{\textit{word} in \textit{GloVe}}
    \STATE word\_vector = get\_vector(\textit{GloVe}, \textit{word})
  \ELSIF{\textit{word} in \textit{Emoji2Vec}}
    \STATE word\_vector = get\_vector(\textit{Emoji2Vec}, \textit{word})
  \ELSE
    \STATE /*$n =$ \textit{Number of characters in word}*/
    \STATE word\_vector = $\frac{1}{n}$ * $\sum_{1}^{n}$ get\_vector(CharEmbed, chars[n])
  \ENDIF
  \ENDFOR
\end{algorithmic}
\caption{Procedure to generate representations}
\label{algo}
\end{algorithm}

\section{Experiments and Results}
\label{sec:exp}
\tikzstyle{in-out}=[draw, fill=red!20, text width=15em, outer sep=2, text centered, minimum height=3em, rounded corners]
\tikzstyle{in-out-large}=[draw, fill=green!30, text width=8em, outer sep=2, text centered, minimum height=5em, rounded corners]
\tikzstyle{process}=[draw, outer sep=2, fill=teal!20, text width=31.5em, text centered, minimum height=3em]
\tikzstyle{process-2}=[draw, outer sep=2, fill=green!30, text width=31.5em, text centered, minimum height=3em, rounded corners]
\tikzstyle{process-small}=[draw, outer sep=2, fill=blue!20, text width=15em, text centered, minimum height=2.5em]
\tikzstyle{process-large}=[draw, outer sep=2, fill=blue!20, text width=31.5em, text centered, minimum height=4em]
\tikzstyle{process-huge}=[draw, outer sep=2, fill=blue!20, text width=38em, text centered, minimum height=6em]

\tikzstyle{in-out-new}=[draw, fill=red!20, text width=5em, outer sep=2, text centered, minimum height=3em, rounded corners]
\tikzstyle{process-2-new}=[draw, outer sep=2, fill=green!30, text width=12em, text centered, minimum height=3em, rounded corners]
\tikzstyle{process-small-new}=[draw, outer sep=2, fill=blue!20, text width=5em, text centered, minimum height=2.5em]
\tikzstyle{process-large-new}=[draw, outer sep=2, fill=blue!20, text width=12em, text centered, minimum height=4em]
\tikzstyle{process-new}=[draw, outer sep=2, fill=teal!20, text width=41.0em, text centered, minimum height=3em]

\tikzstyle{in-out-dummy}=[text width=3em, outer sep=0, text centered, minimum height=3em, rounded corners]

\begin{figure}[ht!]
	\centering
	    \subfloat[Sentiment class distribution. \label{fig:data:class}]
 	    {%
            \resizebox{0.4\textwidth}{!}
            {
 			    \begin{tikzpicture}
                    \begin{axis}[
                        name=plot1,
                        ybar,
                        enlargelimits=0.15,
                        width=\textwidth,
                        height=8.5cm,
                        bar width=0.2cm,
                        ylabel ={Tweets},
                        ylabel near ticks,
                        xlabel={Sentiment},
                        symbolic x coords={Neg-V,Neg-M,Neg-S,Neu,Pos-S,Pos-M,Pos-V},
                        xtick=data,
                        area legend,
                        legend style={font=\scriptsize, legend pos= north east, legend columns=1,  legend style={row sep=0.5pt}},
                        ]
                        \addplot[color=red, fill=red!50,] coordinates {(Neg-V,129) (Neg-M,249) (Neg-S,78) (Neu,341) (Pos-S,167) (Pos-M,92) (Pos-V,125)};\addlegendentry{Train - 1181}
                        \addplot[color=cyan, fill=cyan!50,] coordinates {(Neg-V,69) (Neg-M,95) (Neg-S,34) (Neu,105) (Pos-S,58) (Pos-M,35) (Pos-V,53)};\addlegendentry{Development - 449}
                        \addplot[color=blue, fill=blue!50,] coordinates {(Neg-V,93) (Neg-M,167) (Neg-S,80) (Neu,262) (Pos-S,107) (Pos-M,91) (Pos-V,137)};\addlegendentry{Test - 937}
                    \end{axis}
                \end{tikzpicture}
     	    }
        }
        \hspace{2em}
        \subfloat[Sentiment intensity distribution. \label{fig:data:int}]
    	{%
        \centering
        \includegraphics[width=0.4\textwidth]{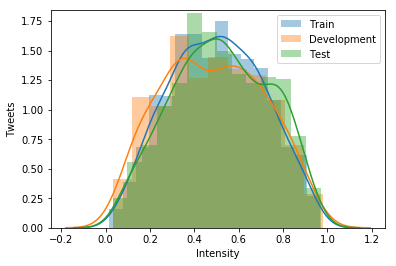}
        }
        \caption{Sentiment distribution for SemEval-2018 task on Affect in Tweets \cite{mohammad2018semeval}}
        \label{fig:data}
\end{figure}
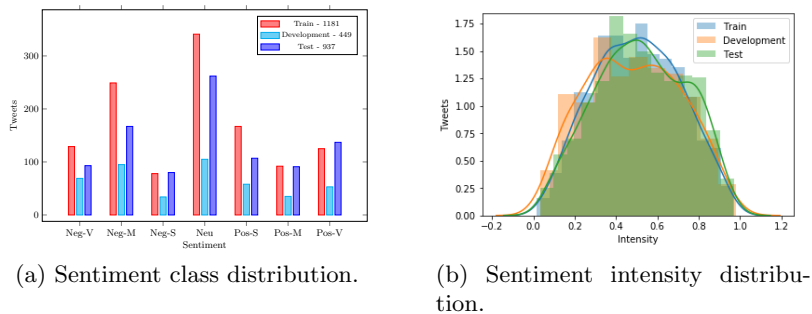
\subsection{Dataset}
We evaluate our proposed model on the datasets of SemEval-2018 shared task on Affect in Tweets \cite{mohammad2018semeval}. There are approximately 1181, 449 \& 937 tweets for training, development and testing. For each tweet, two labels are given: a) sentiment class (one of the seven class on sentiment scale i.e. very positive (\textbf{\em Pos-V}), moderately positive (\textbf{\em Pos-M}), slightly positive (\textbf{\em Pos-S}), neutral (\textbf{\em Neu}), slightly negative (\textbf{\em Neg-S}), moderately negative (\textbf{\em Neg-M}), and very negative (\textbf{\em Neg-V})); and b) an intensity score in the range 0 to 1. We treat prediction of these two labels as two separate tasks. In our multi-task learning framework, we intend to solve these two tasks together. A brief statistics of the datasets is depicted in Figure \ref{fig:data}.

\subsection{Preprocessing}
Tweets in raw form are noisy because of the use of irregular, short form of text (e.g. \textit{hlo}, \textit{whtsgoin} etc.), emojis and slangs and are prone to many distortions in terms of semantic and syntactic structures. The preprocessing step modifies the raw tweets to prepare for feature extraction. We use Ekphrasis tool \cite{baziotis-pelekis-doulkeridis:2017:SemEval2} for tokenization, word normalization, word segmentation (for splitting hashtags) and spell correction. Ekphrasis is a text processing tool, geared towards text from social networks, such as Twitter or Facebook. They used word statistics from 2 big corpora i.e. English Wikipedia and Twitter (330 million English tweets). Ekphrasis was developed as a part of text processing pipeline for SemEval-2017 shared task on Sentiment Analysis in Twitter \cite{rosenthal-farra-nakov:2017:SemEval}. We list the preprocessing steps that have been carried out below. 
\begin{itemize}
    \item All characters in text are converted to lower case
    \item Remove punctuation except ! and ? because '!' and '?' may contribute to better result of valence detection.
    \item Remove extra space and newline character
    \item Group similar emoji, replace them with their meaning in words using Emojipedia
    \item Named Entity recognition and replace with keyword or token (\textit{@shikhar} $\rightarrow$ \textit{username}, \textit{https://www.iitp.ac.in} $\rightarrow$ \textit{url} )
    \item Split the hashtags (\textit{\#iamcool} $\rightarrow$ \textit{i am cool})
    \item Correct the misspelled words (\textit{facbok} $\rightarrow$ \textit{facebook})

\end{itemize}

\subsection{Experiments}
We pad each tweet to a maximum length of 50 words. We employ 300-dimensional word embedding for the experiments (c.f. Section \ref{sec:we}). The GRU dimension is set to 256. We use 100 different filters of varying sizes (i.e. 2-gram, 3-gram, 4-gram, 5-gram and 6-gram filters) with max-pool layer in the CNN module. We use \textit{ReLU} \cite{nair2010rectified} activation and set the \textit{Dropout} \cite{dropout} as 0.5 . We optimize our model using \textit{Adam} \cite{kingma2014adam} with cross-entropy and \textit{mean-squared-error (MSE)} loss functions for sentiment classification and intensity prediction, respectively.

For experiments, we employ python based deep learning library Keras with TensorFlow as the backend. We adopt the official evaluation metric of SemEval-2018 shared task on Affect in Tweets \cite{mohammad2018semeval}, i.e. Pearson correlation coefficient, for measuring the performance of both tasks. We train our model for the maximum 100 epochs with early stopping criteria having patience=20.

\begin{table}[t]
    \caption{Pearson correlation for STL and MTL frameworks for sentiment classification and intensity prediction. $^+$Reported in \cite{duppada2018seernet}; $^*$Reproduced by us.}
    \centering
    \begin{tabular}{|l|c|c||c|c|}
    \hline
        \multirow{2}{*}{\bf Framework} & \multicolumn{2}{c||}{\bf Sentiment Classification} & \multicolumn{2}{c|}{\bf Intensity Prediction} \\ \cline{2-5}
            & DL (Softmax) & ML (SVM) & DL (Sigmoid) & ML (SVR) \\ \hline \hline
        Single-task Learning (STL) & 0.361 & 0.745 & 0.821 & 0.818 \\ \hline
        Multi-task Learning (MTL) & \bf 0.408 & \bf 0.772 & \bf 0.825 & \bf 0.830 \\ \hline \hline
        State-of-the-art \cite{duppada2018seernet} & \multicolumn{2}{c||}{\bf 0.836$^+$ (0.776$^*$)} & \multicolumn{2}{c|}{\bf 0.873$^+$ (0.829$^*$)} \\ \hline
    \end{tabular}
    \label{tab:my_label}
\end{table}

In single-task learning (STL) framework, we build separate systems for both sentiment classification and intensity prediction. We pass the normalized tweet to our Convolutional-GRU framework for learning. Since the number of training samples are considerably few to effectively learn a deep learning model, we assist the model with various hand-crafted features. The concatenated representations are fed to the softmax layer (or sigmoid) for the sentiment (intensity) prediction. We obtain 0.361 Pearson coefficient for sentiment classification and 0.821 for intensity prediction. Further, we also try to exploit the traditional machine learning algorithms for prediction. We extract the concatenated representations and feed them as an input to SVM for sentiment classification and SVR for intensity prediction. Consequently, SVM reports increased Pearson score of 0.745 for sentiment classification, whereas we observe comparable results (i.e. 0.818 Pearson score) for intensity prediction. 

The MTL framework yields an improved performance for both the tasks in both the scenarios. In the first model, MTL reports 0.408 and 0.825 Pearson scores as compared with the Pearson scores of 0.361 \& 0.821 in STL framework for the sentiment classification and intensity prediction, respectively. Similarly, the MTL framework reports 3 and 2 points improved Pearson scores in the second model for the two tasks, respectively. These improvements clearly suggest that the MTL framework, indeed, exploit the inter-relatedness of multiple tasks in order to enhance the individual performance through a joint-model. Further, we observe the improvement of MTL models to be statistically significant with 95\% confidence i.e. \textit{p}-value $<0.05$ for paired \textit{T-test}. 

On same dataset, Duppada et al. \cite{duppada2018seernet} (winning system of SemEval-2018 task on Affect in Tweets \cite{mohammad2018semeval}) reports Pearson scores of 0.836 and 0.873 for sentiment classification and intensity prediction, respectively. The authors in \cite{duppada2018seernet} passed the same handcrafted features individually through XGBost and Random Forest classifier/regressor and combined the results of all the classifiers/regressors using stacking ensemble technique.  After that they passed the results from the models to a meta classifier/regressor as input. They used Ordinal Logistic Classifier and Ridge Regressor as meta classifier/regressor. In comparison, our proposed system (i.e. MTL for ML framework) obtains Pearson scores of 0.772 and 0.830 for sentiment classification and intensity prediction, respectively. It should be noted that we tried to reproduce the works of Duppada et al. \cite{duppada2018seernet}, but obtained Pearson scores of only 0.776 and 0.829, respectively. Further, our proposed MTL model offers lesser complexity compared to the state-of-the-art systems. Unlike the state-of-the-art systems we do not require separate system for each task, rather an end-to-end single model addresses both the tasks simultaneously. 

\subsection{Error Analysis}
\begin{figure}[ht!]
    \centering
    \subfloat[Multi-task:DL]{\includegraphics[width=0.40\textwidth]{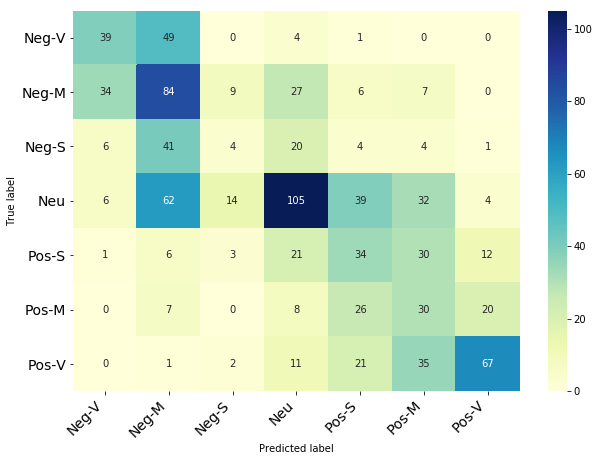}\label{fig:confusion:mtl:dl}}
    \hspace{0.5cm}
     \subfloat[Multi-task:ML]
     {\includegraphics[width=0.40\textwidth]{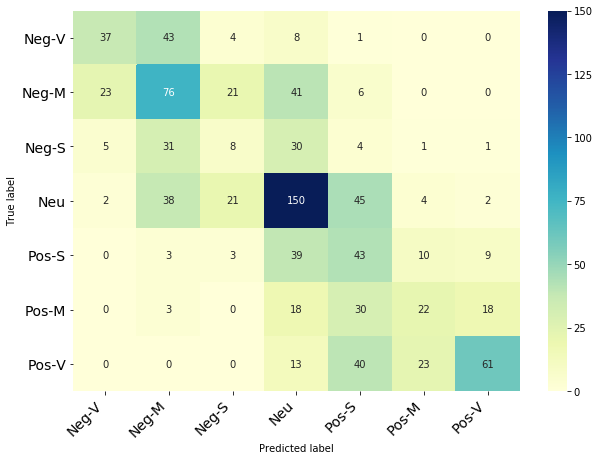}\label{fig:confusion:mtl:ml}}
     \caption{Confusion matrices for sentiment classification.}
     \label{fig:confusion}
\end{figure}

In Figure \ref{fig:confusion}, we present the confusion matrices for both the models (first and second, based on DL and ML paradigms). It is evident from the confusion matrices that most of the mis-classifications are within the close proximity of the actual labels, and our systems occasionally confuse with `\textit{positive}' and `\textit{negative}' polarities (i.e. only 43 and 22 mis-classifications for the first model-DL based and second model-ML based, respectively).

\begin{table}[]
\centering

\caption{MTL vs STL for Sentiment Classification and Intensity Prediction}
\adjustbox{max width=\textwidth}{
\begin{tabular}{lllllll}

\hline

\multicolumn{1}{|l|}{\multirow{2}{*}{\textbf{Sentence}}}                                                                                                                                        & \multicolumn{1}{l|}{\multirow{2}{*}{\textbf{Actual}}} & \multicolumn{2}{c|}{\textbf{DL}}                                                                                                                                   & \multicolumn{1}{l|}{\multirow{2}{*}{}} & \multicolumn{2}{c|}{\textbf{ML}}                                                                                                                                   \\ \cline{3-4} \cline{6-7} 

\multicolumn{1}{|l|}{}                                                                                                                                                                 & \multicolumn{1}{l|}{}                        & \multicolumn{1}{l|}{\textbf{MTL}}                                                    & \multicolumn{1}{l|}{\textbf{STL}}                                                    & \multicolumn{1}{l|}{}                  & \multicolumn{1}{l|}{\textbf{MTL}}                                                    & \multicolumn{1}{l|}{\textbf{STL}}                                                    \\ \hline\hline

\multicolumn{1}{|l|}{\begin{tabular}[c]{@{}l@{}}\em Maybe he was partly right. THESE emails might\\ \em lead to impeachment and 'lock him up' \#ironic\\ \em \#ImpeachTrump\end{tabular}}          & \multicolumn{1}{l|}{\textbf{\textit{Neg-M}}}                   & \multicolumn{1}{l|}{\textit{Neg-M}}                                                  & \multicolumn{1}{l|}{\textcolor{red}{\textit{Neg-V}}}                                                  & \multicolumn{1}{l|}{}                  & \multicolumn{1}{l|}{\textit{Neg-M}}                                                  & \multicolumn{1}{l|}{\textcolor{red}{\textit{Neg-S}}}                                                  \\ \hline

\multicolumn{1}{|l|}{\begin{tabular}[c]{@{}l@{}}\em \#Laughter strengthens \#relationships. \#Women\\\em are more attracted to someone with the ability to\\\em make them \#laugh.\end{tabular}} & \multicolumn{1}{l|}{\textbf{\textit{\textit{Pos-M}}}}                   & \multicolumn{1}{l|}{\textit{Pos-M}}                                                  & \multicolumn{1}{l|}{\textcolor{red}{\textit{Pos-S}}}                                                  & \multicolumn{1}{l|}{}                  & \multicolumn{1}{l|}{\textit{Pos-M}}                                                  & \multicolumn{1}{l|}{\textcolor{red}{\textit{Pos-S}}}                                                  \\ \hline

\multicolumn{7}{c}{a) Sentiment Classification}                                                                                                                                                                                                                                                                                                                                                                                                                                                                                                                                                    \\     \\ \hline

\multicolumn{1}{|l|}{\begin{tabular}[c]{@{}l@{}}\em I graduated yesterday and already had 8 family\\\em members asking what job I've got now \#nightmare\end{tabular}}                        & \multicolumn{1}{l|}{\hfil\textbf{0.55}}                    & \multicolumn{1}{c|}{\begin{tabular}[c]{@{}c@{}}0.57\\ (+0.02)\end{tabular}} & \multicolumn{1}{c|}{\begin{tabular}[c]{@{}c@{}}0.51\\ (-0.04)\end{tabular}} & \multicolumn{1}{l|}{}                  & \multicolumn{1}{c|}{\begin{tabular}[c]{@{}c@{}}0.59\\ (+0.04)\end{tabular}} & \multicolumn{1}{c|}{\begin{tabular}[c]{@{}c@{}}0.64\\ (+0.09)\end{tabular}} \\ \hline

\multicolumn{1}{|l|}{\begin{tabular}[c]{@{}l@{}}\em @rohandes Lets see how this goes. We falter in SL\\\em and this goes downhill.\end{tabular}}                                              & \multicolumn{1}{l|}{\hfil \textbf{0.49}}                        & \multicolumn{1}{c|}{\begin{tabular}[c]{@{}c@{}}0.48\\ (-0.01)\end{tabular}} & \multicolumn{1}{c|}{\begin{tabular}[c]{@{}c@{}}0.35\\ (-0.14)\end{tabular}} & \multicolumn{1}{l|}{}                  & \multicolumn{1}{c|}{\begin{tabular}[c]{@{}c@{}}0.49\\ (+0.00)\end{tabular}} & \multicolumn{1}{c|}{\begin{tabular}[c]{@{}c@{}}0.29\\ (-0.20)\end{tabular}} \\ \hline

\multicolumn{1}{|l|}{\begin{tabular}[c]{@{}l@{}}\em It's kind of shocking how amazing your rodeo fam-\\\em ily is when the time comes that you need someone\end{tabular}}                     & \multicolumn{1}{l|}{\hfil\textbf{0.52}}                        & \multicolumn{1}{c|}{\begin{tabular}[c]{@{}c@{}}0.53\\ (+0.01)\end{tabular}} & \multicolumn{1}{c|}{\begin{tabular}[c]{@{}c@{}}0.55\\ (+0.03)\end{tabular}} & \multicolumn{1}{l|}{}                  & \multicolumn{1}{c|}{\begin{tabular}[c]{@{}c@{}}0.52\\ (+0.00)\end{tabular}} & \multicolumn{1}{c|}{\begin{tabular}[c]{@{}c@{}}0.51\\ (-0.01)\end{tabular}} \\ \hline

\multicolumn{7}{c}{b) Intensity Prediction}                                                                                                                                                                                                                                                                                                                                                                                                                                                                                                                                                           

\end{tabular}
}
\end{table}

We also perform error analysis on the obtained results. Few frequently occurring error cases are presented below:
\begin{itemize}
    \item \textbf {Metaphoric expressions:} Presence of metaphoric/ironic/sarcastic expressions in the tweets makes it challenging for the systems in correct predictions. 
\begin{itemize}
    \item ``{\em @user But you have a lot of time for tweeting \textbf{\#ironic}".} \\ 
    \textbf{Actual:} {\em Neg-M} \qquad\qquad\qquad\qquad\qquad \textbf{Prediction:} {\em Neu}
\end{itemize}
\item \textbf{Neutralizing effect of opposing words:} Presence of opposing phrases in a sentence neutralizes the effect of actual sentiments.
\begin{itemize}
    \item ``{\em @user Macron  slips  up  and  has  a  moment  of  clarity \& common sense... now he is a \textbf{raging racist}. Sounds right. \textbf{Liberal} logic}" \\ 
    \textbf{Actual:} {\em Neg-M} \qquad\qquad\qquad\qquad\qquad \textbf{Prediction:} {\em Neu}
\end{itemize}
\end{itemize}

We further analyze the predictions of our MTL models against STL models. Analysis suggests that our MTL model indeed improves the predictions of many examples that are mis-classified (or having larger error margins) than the STL models. In Table \ref{tab:mtl:stl}, we list a few examples showing the actual labels, MTL prediction and STL prediction for both sentiment classification and intensity prediction.

\section{Conclusion}\label{sec:con}
In this paper, we have presented a hybrid multi-task learning framework for affective language. We propose a convolutional-GRU network with the assistance of a diverse hand-crafted feature set for learning the shared hidden representations for multiple tasks. The learned representation is fed to SVM/SVR classifier for the predictions. We have evaluated our model on the benchmark datasets of SemEval-2018 shared on Affect in Tweets for the two tasks (i.e. sentiment classiffication and intensity prediction). Evaluation suggests that a single multi-task model obtains improved results against separate systems of single-task models.

\section{Acknowledgement}
Asif Ekbal acknowledges the Young Faculty Research Fellowship (YFRF), supported by Visvesvaraya PhD scheme for Electronics and IT, Ministry of Electronics and Information Technology (MeitY), Government of India, being implemented by Digital India Corporation (formerly Media Lab Asia).

\bibliographystyle{splncs04}
\bibliography{mybibliography}

\end{document}